\documentclass[letterpaper, 10 pt, conference]{ieeeconf}  

\IEEEoverridecommandlockouts                              

\usepackage{amsmath, amssymb , graphicx, subcaption}
\usepackage{multirow}
\usepackage{dblfloatfix} 
\usepackage{balance}
\usepackage{hyperref}

\DeclareCaptionFont{caption}{\fontsize{9}{9.6}\selectfont}
\title{\LARGE \bf  Open-Sourced Reinforcement Learning Environments\\ for Surgical Robotics}

\author{Florian Richter$^1$ \IEEEmembership{Student Member, IEEE}, Ryan K. Orosco$^2$ \IEEEmembership{Member, IEEE}, and \\Michael C. Yip$^1$ \IEEEmembership{Member, IEEE}
\thanks{$^1$Florian Richter and Michael C. Yip are with the Department of Electrical and Computer Engineering, University of California San Diego, La Jolla, CA 92093 USA. {\tt\small \{frichter, yip\}@ucsd.edu}}%
\thanks{$^2$Ryan K. Orosco is with the Department of Surgery - Division of Head and Neck Surgery, University of California San Diego, La Jolla, CA 92093 USA. {\tt\small rorosco@ucsd.edu}}}

\begin{document}

\maketitle
\thispagestyle{empty}
\pagestyle{empty}

\begin{abstract}
Reinforcement Learning (RL) is a machine learning framework for artificially intelligent systems to solve a variety of complex problems. 
Recent years has seen a surge of successes solving challenging games and smaller domain problems, including simple though non-specific robotic manipulation and grasping tasks. 
Rapid successes in RL have come in part due to the strong collaborative effort by the RL community to work on common, open-sourced environment simulators such as OpenAI's Gym that allow for expedited development and valid comparisons between different, state-of-art strategies. 
In this paper, we aim to start the bridge between the RL and the surgical robotics communities by presenting the first open-sourced reinforcement learning environments for surgical robots, called dVRL\footnote[3]{dVRL available at  \url{https://github.com/ucsdarclab/dVRL}}.
Through the proposed RL environments, which are functionally equivalent to Gym, we show that it is easy to prototype and implement state-of-art RL algorithms on surgical robotics problems that aim to introduce autonomous robotic precision and accuracy to assisting, collaborative, or repetitive tasks during surgery. 
Learned policies are furthermore successfully transferable to a real robot. 
Finally, combining dVRL with the over 40+ international network of da Vinci Surgical Research Kits in active use at academic institutions, we see dVRL as enabling the broad surgical robotics community to fully leverage the newest strategies in reinforcement learning, and for reinforcement learning scientists with no knowledge of surgical robotics to test and develop new algorithms that can solve the real-world, high-impact challenges in autonomous surgery.
\end{abstract}

\section{Introduction}

Reinforcement Learning (RL) is a framework that has been utilized in areas largely outside of surgical robotics to incorporate artificial intelligence to a variety of problems \cite{sutton_book_rl}. The problems solved, however, have mostly been in extremely structured environments such as video games \cite{QLearning_1_atari} and board games \cite{alphaGo}. There has also been recent success in robotic manipulation and specifically grasping, and with evidence that the learned policies are transferable from simulation to real robots \cite{domain_random_1, domain_random_2}. These successes have hinged on having simulation environments that are lightweight and efficient, as RL tends to require thousands to millions of simulated attempts to evaluate and explore policy options. For robotics, this is crucial for real-world use of RL due to the impracticality of running millions of attempts on a physical system only to learn a low-level behavior. 


\begin{figure}[t]
	\centering
	\vspace{2mm}
  	\includegraphics[width=\linewidth]{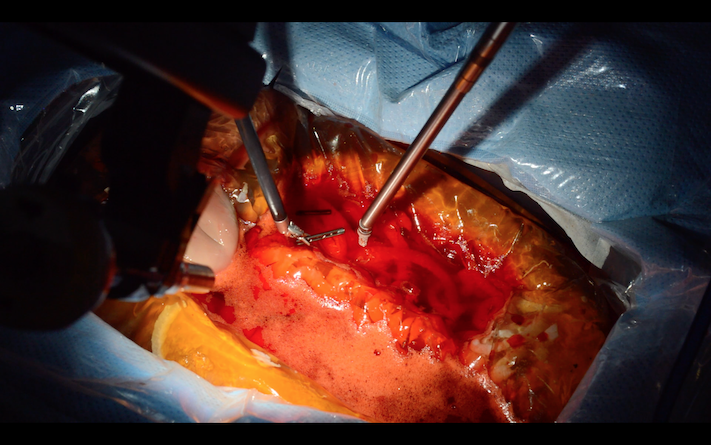}
    \caption{Reinforcement Learning in Action: we used a learned policy from our RL environment in a a collaborative human-robot context, perform autonomous suction (right arm) of blood to iteratively reveal several debris that a surgeon-controlled arm then removes from a simulated abdomen.}
    \label{fig:firstpic}
\end{figure}

Surgical robots, such as Intuitive Surgical's da Vinci\textregistered{} Surgical System, have brought about more efficient surgeries by improving the dexterity and reducing fatigue of the surgeon through teleoperational control. While these systems are already providing great care to patients, they have also opened the door to a variety of research including surgeon performance metrics \cite{dVLogger}, remote teleoperation \cite{teleop_delay_1}, and surgical task automation \cite{Nikhil}.  Surgical task automation have furthermore been an increasing area of research in an effort to improve patient throughput, reduce quality-of-care variance among surgeries, and potentially deliver automated surgery in the future. Automation efforts includes automating subtasks includes knot tying \cite{knot_tying_1, knot_tying_2}, compliant object manipulation \cite{alambeigi2018robust}, endoscopic motions \cite{ecm_motion_1}, surgical cutting \cite{rl_cutting, autonomous_cutting_1}, suture needle manipulation \cite{suture_needle_pick_up, zhong2019dual} and debris removal \cite{debris_removal_1,  debris_removal_3}. One of the challenges moving forward for the surgical robotics community is that despite these successes, many have been based around hand-crafted control policies that can be difficult to both develop at scale and generalize across a variety of environments. RL offers a solution to these problems by shifting human time-costs and the limitations of feature- and controller-design, to autonomously learning these via large-scale, faster-than-real-time, parallelized simulations  (Fig. \ref{fig:firstpic}).

To bridge reinforcement learning with surgical robotics, simulation environments need to be provided such that RL algorithms of past, present, and future can be prototyped and tested on. OpenAI's Gym \cite{openAI_gym} has offered perhaps one of the most impactful resource to the RL community for testing a range of environments and domains through a common API, and has been wildly successful in engaging a broad range of machine learning researchers, engineers, and hobbyists. This is primarily due to its incredibly simple interface, with mainly four function calls (\texttt{make, reset, step, } and \texttt{render}) that allows all kinds of scenarios to be learned. In this paper, we aim to bring RL to the surgical robotics domain via the first open-sourced reinforcement learning environments for the da Vinci Research Kit (dVRK) \cite{DVRK}, called dVRL. We are motivated to engage the broader community that include surgical robotics and also non-domain experts, such that RL enthusiasts with no domain knowledge of surgery can still easily prototype their algorithms with such an environment and contribute to solutions that would have real world significance to robotic surgery and the patients that undergo those procedures. To move towards this goal, we present the following novel contributions:
\begin{enumerate}
    \item the first, open-sourced reinforcement learning environment for surgical robotics and
    \item demonstration of learned policies from the RL environment effectively transferring to a real robot with minimal effort.
\end{enumerate}
The syntactic interface with the environment is inherited from OpenAI's Gym environment \cite{openAI_gym} and its simple interfaces, and is thus easy to include into their pipeline of environments to test. The RL environments are developed for the widely used dVRK such that any RL-learned strategy could be applied on their platforms. Specifically, newly learned policies can be transferred onto any of the internationally networked, 40+ da Vinci Research Platforms and participating labs \cite{dvrk_wiki}, including the one at UC San Diego, to encourage international collaborations and reduce the barriers for all to validate on a real world system. 

\section{Background in RL}
The RL framework considered is based on a Markov Decision Process where an agent interacts with an environment. The environment observations are defined by the state space $\mathcal{S}$ and the agent interacts with the environments through the action space $\mathcal{A}$. The initial state is sampled from a distribution of initial states $\mathbb{P}(S_0 = s_0)$ where $s_0 \in \mathcal{S}$. When an agent performs an action, $a_t \in \mathcal{A}$, on the environment, the next state is sampled from the transition probability $\mathbb{P}(S' = s_{t+1}| S = s_t, A = a_t)$ where $s_t, s_{t+1} \in \mathcal{S}$ and a reward $r_t$ is generated from a reward function $r: \mathcal{S} \times \mathcal{A} \rightarrow \mathbb{R}$.

In RL, the agent aims to find a policy $\pi: \mathcal{S} \rightarrow \mathcal{A}$ that maximize the cumulative reward, $G_t = \sum_{i=t}^{T+t} \gamma^{i-t}r_i$ where $T$ is the time horizon and $\gamma \in [0,1]$ is the discount factor. The Q-Function, $Q^{\pi}(s_t,a_t) = \mathbb{E}_\pi [ G_t| S=s_t, A=a_t]$, gives the expected value of the cumulative reward when in state $s_t$, taking an action $a_t$, and following the policy $\pi$. Therefore an optimal policy for an agent $\pi^*$, which aims to maximizes the cumulative reward, can be formalized as $Q^{\pi^*}(s_t,a_t) \geq Q^{\pi}(s_t,a_t)$ for all $s_t \in \mathcal{S}$, $a_t \in \mathcal{A}$, and policies $\pi$. $Q^{\pi^*}(s_t,a_t)$ is considered the optimal Q-Function.

There is a substantial amount of research in RL to find the optimal policy. A few examples are: policy gradient methods, which solve for the policy directly \cite{policyGrad_1, policyGrad_2}, Q-Learning that solve for the optimal Q-Function \cite{QLearning_1_atari, QLearning_2}, and actor-critic methods which find both \cite{actorCritic, ddpg}. OpenAI also created a well established standard in the RL community for developing new environments to allow for easier evaluation of RL algorithms \cite{openAI_gym}. By creating syntatic parallels, the state-of-art in RL may be directly applied to surgical robot platforms via dVRL.

\section{Methods}
The environments presented inherit from the OpenAI Gym Environments and utilize the V-REP physics simulator developed by Fontanelli et al. \cite{vrep_simulator}. 
V-REP was chosen due to its recent success in other deep learning applications for robotic control \cite{james2019pyrep} and easy usage with environment creation, various sensors, and thread simulation \cite{thread_simulation}.
When instantiated, the simulated environment is created and communicated through V-REP's remote API in synchronous mode. To ensure safe creation and deletion of the simulated environment, the V-REP simulation is ran in a separate docker container. This also allows multiple instances of the environments in the same system, which can be utilized for distributed RL \cite{distributedRL_1}.

\subsection{Simulation Details}

\begin{figure}[b]
	\centering
	\vspace{2mm}
	\includegraphics[width=0.48\textwidth]{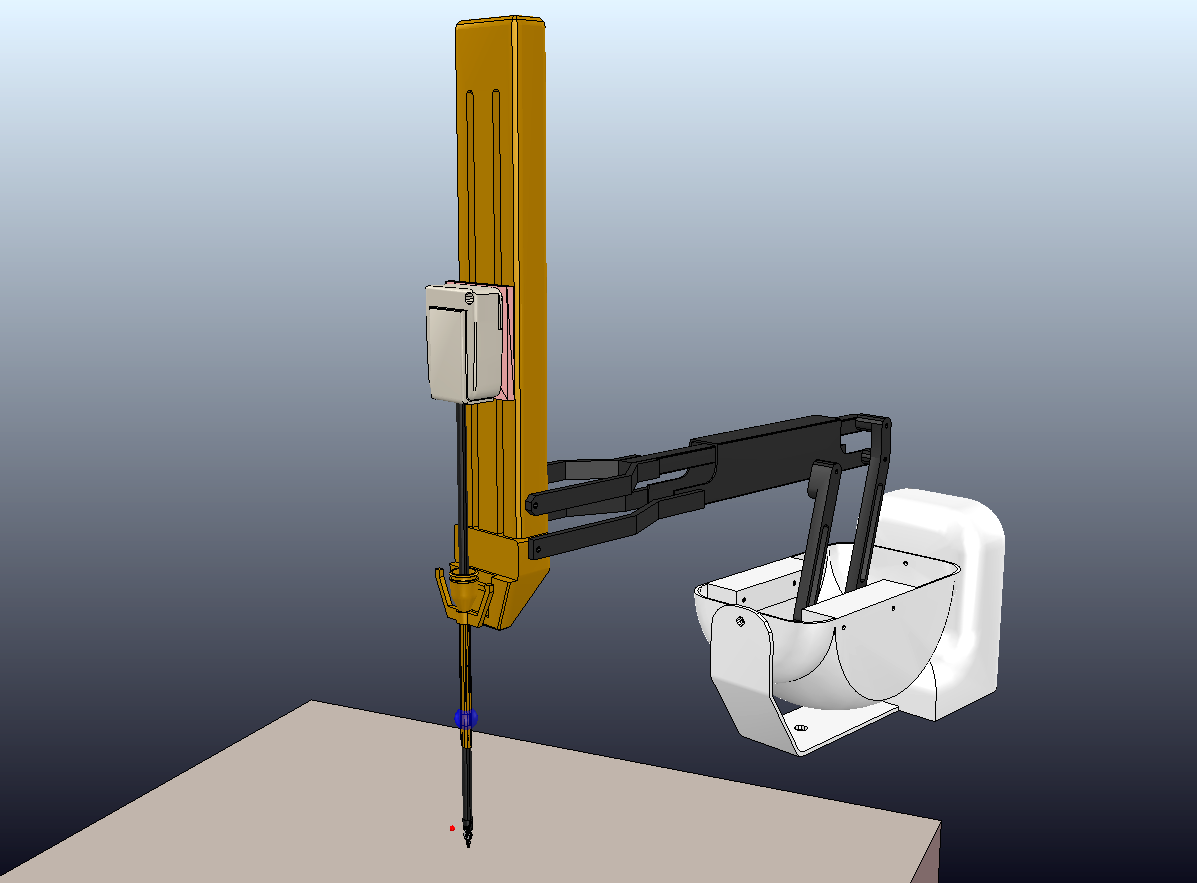}
	\caption{Simulation scene in V-REP of the single PSM arm. This is the fundamental scene that the presented environments, PSM Reach and PSM Pick, are based on.}
	\label{fig:psm_arm}
\end{figure}

\begin{figure*}[b]
	\centering
	\vspace{2mm}
	\begin{subfigure}{.24\textwidth}
  		\centering
  		\includegraphics[width=1\linewidth]{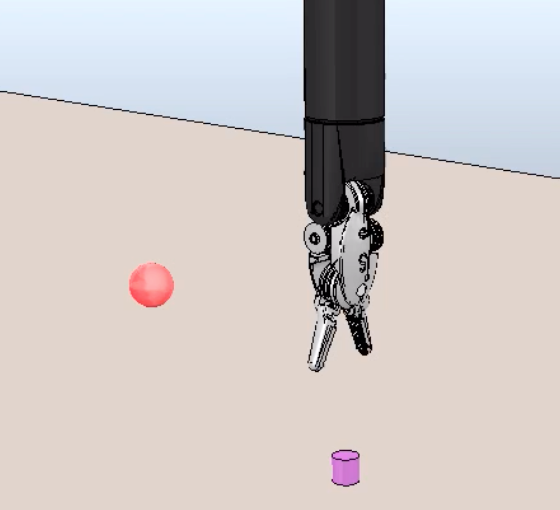}
	\end{subfigure}
	\hspace{0.06mm}
	\begin{subfigure}{.24\textwidth}
		\centering
  		\includegraphics[width=1\linewidth]{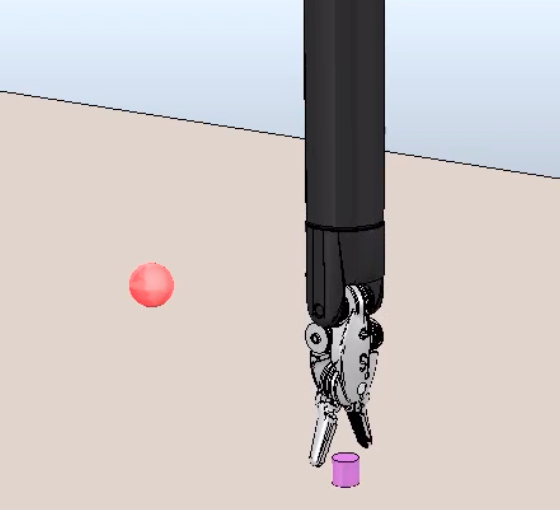}
	\end{subfigure}
	\hspace{0.06mm}
	\begin{subfigure}{.24\textwidth}
  		\centering
  		\includegraphics[width=1\linewidth]{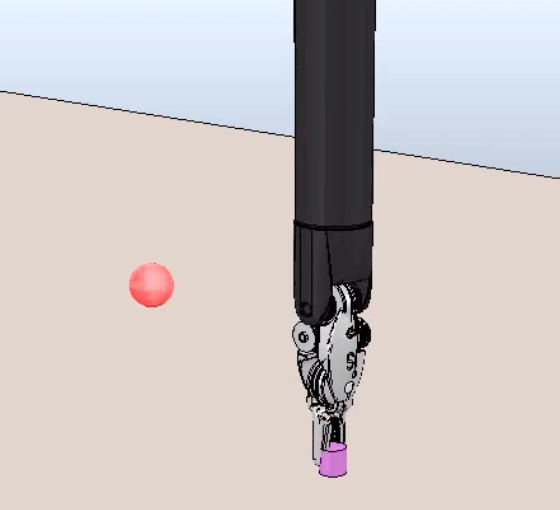}
	\end{subfigure}
	\hspace{0.06mm}
	\begin{subfigure}{.23\textwidth}
		\centering
  		\includegraphics[width=1\linewidth]{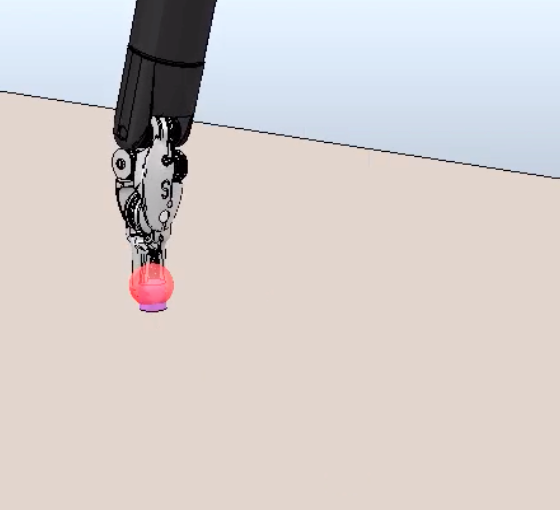}
	\end{subfigure}
	\caption{Example policy solving the PSM Pick Environment. The purple cylinder is the object, and the red sphere is the goal. From left to right the following is done: move to the object, grasp the object, transport the object to the goal.}
	\label{fig:pick_task}
\end{figure*}

The presented environments only utilize one slave arm from the dVRK as shown in Fig. \ref{fig:psm_arm}, also known as a Patient Side Manipulator (PSM) arm. New environments can be easily scaled through the addition of multiple PSM arms and the endoscopic camera arm. The PSM arms on dVRK also have a variety of attachable tools, known as EndoWrists, to accomplish different surgical tasks. The current environments use the Large Needle Driver (LND), which has a jaw gripper to grab objects such as suturing needles. Other tools can be supported in simulation by switching out the tool portion of the model in V-REP. 

The environments also work in the end-effector space rather than the joint space so trained policies that do not require specific tooling, such as the gripper, can transfer to the real dVRK for a variety of tools since each tool has unique kinematics. Furthermore, end-effector control is how surgeons operate the da Vinci\textregistered{} Surgical System. This gives the flexibility to use demonstrations from real operations. For the sake of simplicity, the end-effector orientation is held constant. Therefore, the PSM can be characterized by the three dimensional end-effector position $\mathbf{p}_t$ in its base frame and jaw angle $j_t$. 

To set the workspace for the environments, it is bounded by range $\rho > 0$ and centered around position $\mathbf{p}_c$. So the workspace can be written as:

\begin{equation}
    [\mathbf{p}_c]_i -  \rho \leq [\mathbf{p}_t]_i \leq [\mathbf{p}_c]_i + \rho
\end{equation}
where $i = 1,2,3$ and $[\text{ }\cdot{}\text{ }]_i$ is the i-th dimension of the vector. In addition, the workspace is limited by the joint limits of the PSM arm and obstacles in the environment. Currently, a table is the only obstacle, but more obstacles can be added.

The jaw angle is bounded inclusively from 0 to 1, where 0 is completely closed and 1 is completely open. The values $j_t$ takes on directly correlate with the values used on the real LND during operation. 

To grasp an object in simulation, there is a proximity sensor placed in the gripper of the LND. The object is considered rigidly attached to the gripper if the jaw angle is less than 0.25 and the proximity sensor is triggered. In one of the presented environments, there is a single, small cylindrical object and only its three dimensional position in the PSM arm base frame, $\mathbf{o}_t$, is utilized in the state space.

Due to the millimeter scale the PSM arms operate at, the positions are normalized by the range of the environment. Normalization of both states and actions is regularly used by popular RL libraries and performance improvements has been empirically found \cite{baselines, garage}. The normalized end-effector position and object position are:
\begin{align}
    \widetilde{\mathbf{p}}_t &= (\mathbf{p}_t - \mathbf{p}_c)/\rho \label{equ:normalize_1} \\
    \widetilde{\mathbf{o}}_t &= (\mathbf{o}_t - \mathbf{p}_c)/\rho
    \label{equ:normalize_2}
\end{align}
Another advantage of making the states relative to $\mathbf{p}_c$, is that the learned policies can be rolled out to various joint configurations by re-centering the states. 

Since the orientation is fixed and the PSM arms are operated in the end-effector space, the actions change the end-effector position and set the jaw angle directly. This matches the real da Vinci\textregistered{} Surgical System. To keep the actions normalized between -1 and 1, the next state equation for the PSM arm is:
\begin{align}
    \mathbf{p}_{t+1} &= \eta \mathbf{\Delta}_t + \mathbf{p}_t \\
    j_{t+1} &= (\phi_t + 1)/2
\end{align}
where the elements of $\mathbf{\Delta}_t$ and $\phi_t$ are bounded from -1 to 1 and are considered the actions that can be applied to the environment. 
The $\eta$ term is critical to ensuring effective transfer of policies from the simulation to the real robot. 
On the dVRK, joint level control utilized \cite{DVRK}, so every new end-effector position gives new set points for the joint angles through inverse kinematics. 
This means overshoot or even instability can occur if the difference between the new set point and current joint angle is too great. 
By choosing a value for $\eta$ that ensures negligible overshoot and no instability on the real robot, no dynamics are required for the simulation of the PSM arm, which significantly speeds up the simulation time.
Furthermore, prior work has shown the difficulty in modelling the dynamics of the PSM arm \cite{model_dyn, wang2019convex}, and using the dynamics would require a separate model for each real PSM arm.

\subsection{PSM Reach Environment}
The PSM Reach environment is similar to the Fetch Reach environment \cite{fetchEnv}. The environment aims to find a policy to move the PSM arm to a goal position, $\mathbf{g}$, given a starting position $\mathbf{p}_0$. This type of environment is called a goal environment where an agent is capable of accomplishing multiple goals in a single environment \cite{HER}. The state and action space of the environment is:
\begin{align}
    s_t &= \begin{bmatrix} \widetilde{\mathbf{p}}_t & \widetilde{\mathbf{g}} \end{bmatrix} \\
    a_t &= \begin{bmatrix} \mathbf{\Delta}_t \end{bmatrix}
\end{align}
where $\widetilde{\mathbf{g}}$ is normalized in a similar fashion as Equation (\ref{equ:normalize_1}) and (\ref{equ:normalize_2}). When resetting the environment to begin training, $\mathbf{g}$ and $\mathbf{p}_0$ are uniformly sampled from the workspace previously specified. The reward function is:
\begin{equation}
    r(s_t) = \begin{cases} 
                    -1 & \rho||\widetilde{\mathbf{p}}_t - \widetilde{\mathbf{g}}|| > \delta\\
                     0 & \text{otherwise}
                \end{cases}
\end{equation}
where $\delta$ is the threshold distance. By giving a negative reward until it reaches the goal, the policy should learn to also minimize distance to reach the goal. Note that this environment only uses the end-effector position, so the policy can be applied to all EndoWrists.

\subsection{PSM Pick Environment}

The PSM Pick environment is also a goal environment and similar to the Fetch Pick environment \cite{fetchEnv}. The agent needs to reach to the object at $\mathbf{o}_t$ from a starting position $\mathbf{p}_0 = \mathbf{p}_c$, grasp the object, and move the object to the goal position $\mathbf{g}$. This sequence is shown in Fig. \ref{fig:pick_task}. The state space is:
\begin{align}
    s_t &= \begin{bmatrix} \widetilde{\mathbf{p}}_t & 2j_t - 1 & \widetilde{\mathbf{o}}_t & \widetilde{\mathbf{g}}\end{bmatrix} \\
    a_t &= \begin{bmatrix} \mathbf{\Delta}_t & \phi_t \end{bmatrix}
\end{align}
Similar to the PSM Reach environment, $\mathbf{g}$ is uniformly sampled from the workspace when resetting the environment. The starting position of the object $\mathbf{o}_0$ is placed directly below the gripper on the table. The reward function is:
\begin{equation}
    r(s_t) = \begin{cases} 
                    -1 & \rho||\widetilde{\mathbf{o}}_t - \widetilde{\mathbf{g}}|| > \delta\\
                     0 & \text{otherwise}
                \end{cases}
\end{equation}
where $\delta$ is once again the threshold distance. 

\section{Experiments}
To show the efficiency of the simulated environments, performance measurements are made. State of the art RL algorithms are utilized to solve the environments in simulation. The learned polices are then transferred to the dVRK \cite{DVRK} running at 50Hz. The policy transfer is evaluated individually by replicating the simulated scene and completion of the surgical: tasks suction and debris removal. Both the training of the RL policies and dVRK ran on an Intel\textregistered{} Core\texttrademark{} i9-7940X Processor and NVIDIA's GeForce RTX 2080.

\subsection{Solving Environments}
Both the PSM Reach and PSM Pick environments are given 100 steps per episode with no early termination and the threshold, $\delta$, is set to 3mm. The range $\rho$ is set to 5 cm and 2.5 cm for PSM Reach and PSM Pick respectively. Through experimentation on the dVRK, we found $\eta = 1$ mm to be the highest value where the PSM joints do not overshoot at 50Hz.

The environments are solved in simulation using Deep Deterministic Policy Gradients (DDPG) \cite{ddpg}. DDPG is from the class of Actor-Critic algorithms where it approximates both the policy and Q-Function with separate neural networks. The Q-Function is optimized by minimizing the Bellman loss:
\begin{equation}
    \mathcal{L}_Q = (Q(s_t,a_t) - (r_t + \gamma Q(s_{t+1}, a_{t+1}))^2
\end{equation}
and the policy is optimized by minimizing:
\begin{equation}
\mathcal{L}_{\pi} = - \mathbb{E}_{s_t}[Q(s_t, \pi(s_t)]
\end{equation}

Hindsight Experience Replay (HER) is used as well to generate new experiences for faster training \cite{HER}. HER generates new experiences for the optimization of the policy and/or Q-Function where the goal portion of the state is replaced with previously achieved goals. This improves the sample efficiency of the algorithms and combats the challenge of sparse rewards, which is the case for both the PSM Reach and PSM Pick environments.

The size of the state space relative to the distance the maximum action is very large in the presented environments. This makes exploration very challenging, especially for the PSM Pick environment. To overcome this, demonstrations $\{(s^d_i, a^d_i)\}_{i=0}^{N_d}$ which reach the goal, are generated in simulation and the behavioral cloning loss:
\begin{equation}
    \mathcal{L}_{BC} = \sum \limits_{i=0}^{N_d}||\pi(s^d_i) - a^d_i||^2 
\end{equation}
is augmented with the DDPG policy loss as done by Nair et al.
\cite{behavioral_cloning}. OpenAI Baselines implementation and hyper parameters of DDPG + HER, with the addition of the augmented behavioral cloning, was used \cite{baselines}. 

\subsection{Transfer to Real World}
Using the LND tool with dVRK, the policies are tested on the real system after completing training in simulation. The positional state information for the end-effector is found by calculating forward kinematics from encoder readings. The PSM Reach policy transfer is evaluated by giving random goal locations and seeing if the threshold distance to the goal is met. The PSM Pick Environment is rolled out in a recreated scene of the simulation including the initial PSM position, initial object position, and table location. To simplify the recreated scene, the object position is assumed rigidly attached to the end-effector if the jaw is closed, similar to how the object is grasped in simulation, but this time blind. The object in this experiment is a small sponge.

\subsection{Suction \& Irrigation Tool}
The PSM Reach policy can be rolled out on any EndoWrist since it does not use any tool specific action. To show this, both LND and the Suction \& Irrigation EndoWrists were utilized to rollout the PSM Reach policy on the dVRK. The Denavit Hartenberg (DH) parameters for both tools are shown in Table \ref{tab:dh_param}. The table highlights the variability of the kinematics for EndoWrists. Note that $q_i$ for $i=1,...,6$ is the joint configuration, $a$ and $\alpha$ represents positional and rotational change respectively along the x-axis relative to the previous frame, and $D$ and $\theta$ represents positional and rotational change respectively along the z-axis relative to the frame transformed by $a$ and $\alpha$.

\begin{table}[h]
\centering
\caption{\\DH Parameters for LND and Suction \& Irrigation}
\setlength\tabcolsep{0.5em}
\begin{tabular}{c|cccc|cccc}
       & \multicolumn{4}{c|}{\textbf{LND}} & \multicolumn{4}{c}{\textbf{Suction \& Irrigation}} \\ \hline
      Frame & a & $\alpha$ & D & $\theta$ & a & $\alpha$ & D & $\theta$ \\
      1 & 0 & $\frac{\pi}{2}$ & 0 & $q_1$ + $\frac{\pi}{2}$ & 0 & $\frac{\pi}{2}$ & 0 & $q_1 + \frac{\pi}{2}$ \\
      2 & 0 & $-\frac{\pi}{2}$ & 0 & $q_2$ - $\frac{\pi}{2}$ & 0 & $-\frac{\pi}{2}$ & 0 & $q_2 - \frac{\pi}{2}$ \\
      3 & 0 & $\frac{\pi}{2}$ & $q_3 - l_l$ & 0 & 0 & $\frac{\pi}{2}$ & $q_3 - l_2$ & 0 \\
      4 & 0 & 0 & $l_3$ & $q_4$ & - & - & - & - \\
      5 & 0 & $-\frac{\pi}{2}$ & 0 & $q_5 - \frac{\pi}{2}$ & 0 & $-\frac{\pi}{2}$ & 0 & $q_5 - \frac{\pi}{2}$ \\
      6 & $l_4$ & $-\frac{\pi}{2}$ & 0 & $q_6 - \frac{\pi}{2}$ & $l_5$ & $-\frac{\pi}{2}$ & 0 & $q_6 - \frac{\pi}{2}$ 
\end{tabular}
\label{tab:dh_param}
\end{table}

The Suction \& Irrigation tool was integrated into dVRK with slight modifications to the configuration files. 
Furthermore, the analytical inverse kinematics that is used to set the end-effector is:
\begin{align*}
    &\theta_1 = \text{tan}^{-1} \Big( \frac{p_z}{p_x} \Big) \\
    &\theta_6 = \text{cos}^{-1} 
    \big( \text{sin}(\theta_1)v_x - \text{cos}(\theta_1)v_z \big) \\
    &\text{sin}(\theta_2 + \theta_5) = -\frac{v_y}{\text{sin}(\theta_6)} \\
    &\text{cos}(\theta_2 + \theta_5) = -\frac{v_x\text{cos}(\theta_1) + v_y\text{sin}(\theta_1)}{\text{sin}(\theta_6)} \\
    &\theta_2 = \text{tan}^{-1}\Bigg( \frac{ \frac{p_x}{\text{cos}(\theta_1)} - l_5 \text{cos}(\theta_2 + \theta_5)}{-p_y + l_5\text{sin}(\theta_2 + \theta_5)} \Bigg) \\
    &q_3 = \frac{-p_y + l_5 \text{sin}(\theta_2 + \theta_5)}{\text{cos}(\theta_2)} + l_2 \\
    &\theta_5 = \text{tan}^{-1}\Bigg( \frac{\text{sin}(\theta_2 + \theta_5)}{\text{cos}(\theta_2 + \theta_5)} \Bigg) - \theta_2 
\end{align*}
where $\begin{bmatrix} p_x, p_y, p_z \end{bmatrix}^\top$ and $\begin{bmatrix} v_x, v_y, v_z \end{bmatrix}^\top$ are the position and direction of the end-effector respectively and $\theta_i$ refer to the DH parameter. Note that the orientation of the Suction \& Irrigation tool can be defined by a single directional vector since the tool tip is symmetric about the roll axis. 


\subsection{Suction and Debris Removal}


A simulated abdomen was created by molding pig liver, sausage, and pork rinds in gelatin. The gelatin mold has two large cavities that can be filled with fake blood made by food coloring and water. The surgical task is to use the Suction \& Irrigation tool to remove the fake blood and the LND to grasp and hand the debris, revealed by the suction, to the first assistant. The debris used is a 3 mm by 28 mm dowel spring pin. 

The suction tool uses the policy trained by the PSM Reach Environment. The experiment was repeated where the LND is tele-operated by an expert surgeon who regularly gives care with the da Vinci\textregistered{} Surgical System and autonomously controlled by using both PSM Pick and PSM Reach learned policies to grasp the debris and to hand the debris to the first assistant. For the policies, the goal locations are preset by manually moving the arms to the goals and saving the position. The PSM Pick task in the experiment also uses the same simplification as previously described. To bring the LND in position to pick the debris, the learned PSM Reach policy is used.

\section{Results}

\begin{figure}[t]
	\centering
	\vspace{2mm}
	\includegraphics[width=0.48\textwidth]{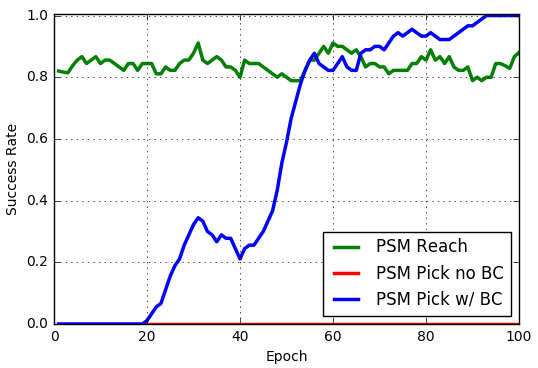}
	\caption{Results of training PSM Reach and Pick using DDPG + HER and Behavioral Cloning (BC). Each epoch is six environments rolling out 50 times per environment for training. The success rate is the average number of times the final state reaches the goal within the threshold from 50 runs.}
	\label{fig:resultsPlot}
\end{figure}

\begin{figure}[b]
	\centering
	\vspace{2mm}
  	\includegraphics[width=0.48\linewidth]{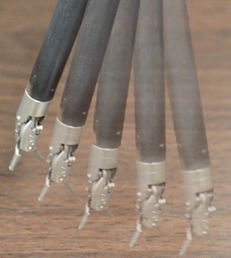}
    \includegraphics[width=0.48\linewidth]{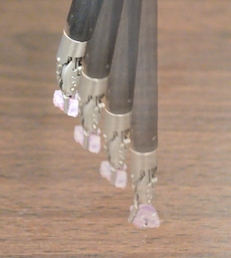}
	\caption{Trained PSM Reach and PSM Pick policies rolled out on the da Vinci Research Kit in the left and right figure respectively. }
	\label{fig:individual_rollout}
\end{figure}

\begin{figure*}[t]
	\centering
	\vspace{2mm}
	\begin{subfigure}{.3\textwidth}
  		\centering
  		\includegraphics[width=1\linewidth]{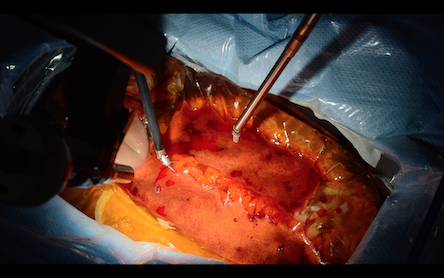}
	\end{subfigure}
	\hspace{1mm}
	\begin{subfigure}{.3\textwidth}
		\centering
  		\includegraphics[width=1\linewidth]{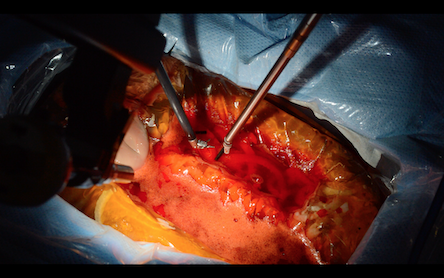}
	\end{subfigure}
	\hspace{1mm}
	\begin{subfigure}{.3\textwidth}
  		\centering
  		\includegraphics[width=1\linewidth]{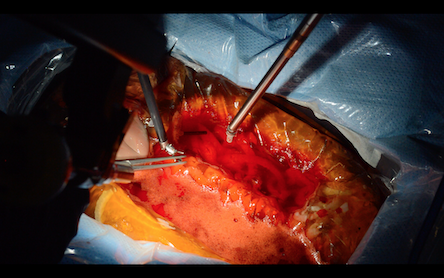}
	\end{subfigure}
	\vskip\baselineskip
	\begin{subfigure}{.3\textwidth}
		\centering
  		\includegraphics[width=1\linewidth]{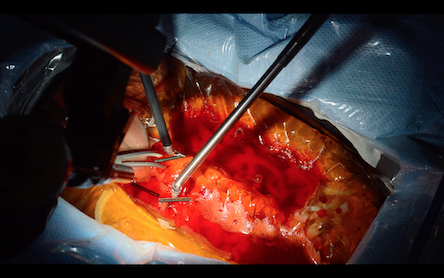}
	\end{subfigure}
	\hspace{1mm}
	\begin{subfigure}{.3\textwidth}
		\centering
  		\includegraphics[width=1\linewidth]{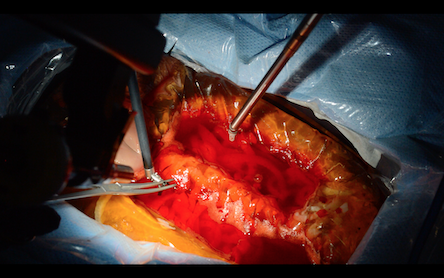}
	\end{subfigure}
	\hspace{1mm}
	\begin{subfigure}{.3\textwidth}
		\centering
  		\includegraphics[width=1\linewidth]{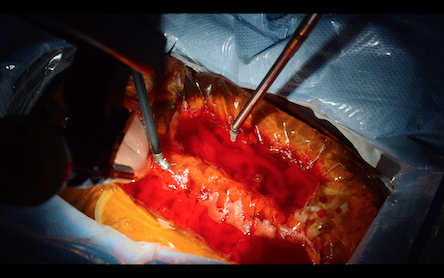}
	\end{subfigure}
	\caption{The suction tool using a trained PSM Reach policy to remove fake blood to reveal debris so the surgeon can remove them from a simulated abdomen. After located and removed by teleoperational control from the simulated abdomen, the debris is handed off to the first assistant.}
	\label{fig:teleop_suction}
\end{figure*}

\begin{figure*}[t]
    \vspace{-4mm}
	\centering
	\begin{subfigure}{.3\textwidth}
  		\centering
  		\includegraphics[width=1\linewidth]{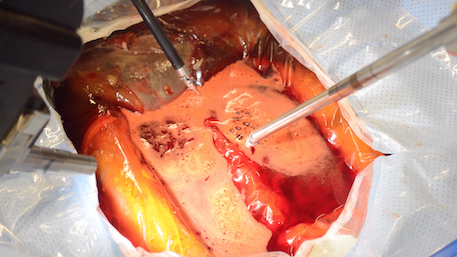}
	\end{subfigure}
	\hspace{1mm}
	\begin{subfigure}{.3\textwidth}
		\centering
  		\includegraphics[width=1\linewidth]{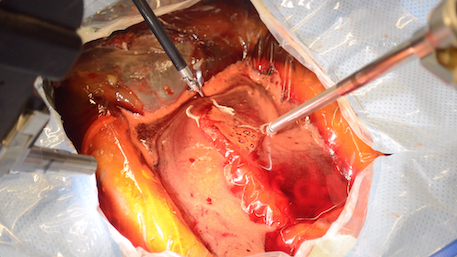}
	\end{subfigure}
	\hspace{1mm}
	\begin{subfigure}{.3\textwidth}
  		\centering
  		\includegraphics[width=1\linewidth]{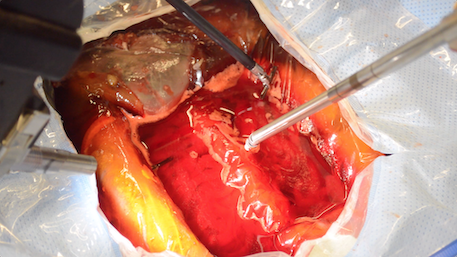}
	\end{subfigure}
	\vskip\baselineskip
	\begin{subfigure}{.3\textwidth}
		\centering
  		\includegraphics[width=1\linewidth]{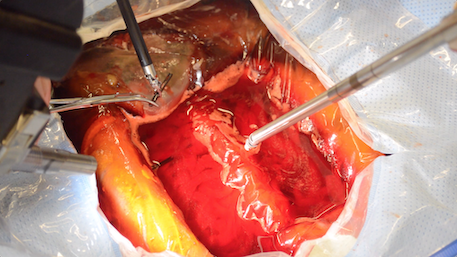}
	\end{subfigure}
	\hspace{1mm}
	\begin{subfigure}{.3\textwidth}
		\centering
  		\includegraphics[width=1\linewidth]{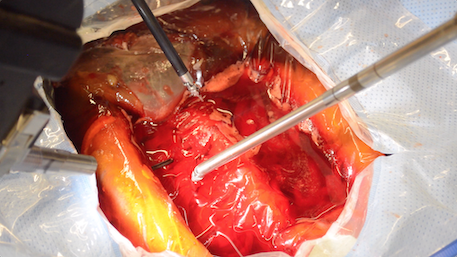}
	\end{subfigure}
	\hspace{1mm}
	\begin{subfigure}{.3\textwidth}
		\centering
  		\includegraphics[width=1\linewidth]{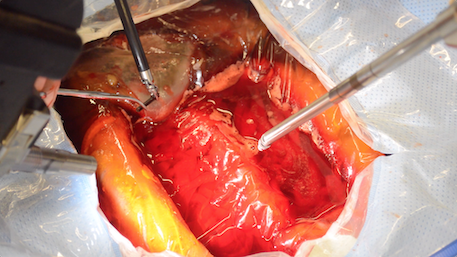}
	\end{subfigure}
	\caption{The suction tool using a trained PSM Reach policy to remove fake blood to reveal debris. After the debris is revealed, the Large  Needle Driver utilized a composition of trained PSM Reach and PSM Pick policies to remove the debris and hand it to the first assistant.}
	\label{fig:autonomous_suction}
\end{figure*}

The timing results of the environments are shown in table \ref{tab:timing_results}. As seen in the table, the parallelization optimization by running the simulations in separate docker containers can allow for more efficient training of RL algorithms. The results from training both PSM Reach and Pick with DDPG + HER are shown in Fig. \ref{fig:resultsPlot}. Note that a rollout is considered successful if the final state gives a reward of 0 which occurs when the goal is reached within the threshold distance. Without behavioral cloning, we were unable to solve the PSM Pick environment. When analyzing the final trained PSM Reach policy, the policy can reach the goal with 100\% success rate if given 1000 simulation steps instead of 100.

\begin{table}[h]
    \centering
    \caption{\\Timing Results of one rollout per Environment}
    \begin{tabular}{c|c|c}
        Num. of Env. & PSM Reach & PSM Pick \\ \hline
         1   & 2.09 sec & 2.09 sec  \\
         2   & 2.36 sec & 2.35 sec  \\
         4   & 2.78 sec & 2.78 sec  \\
         6   & 3.03 sec & 3.02 sec \\
         8   & 3.27 sec & 3.26 sec  \\
    \end{tabular}
    \label{tab:timing_results}
\end{table}

Photos of rolling out the learned PSM Reach and PSM Pick policies are shown in Fig. \ref{fig:individual_rollout}. The policies used were the final PSM Reach policy and the final PSM Pick policy with Behavioral Cloning from training. Both policies were able to reach the threshold distance of 3 mm with 100\% success rate for ten randomly chosen goal locations.

Photos showing the surgical suction and debris removal are in Fig. \ref{fig:teleop_suction} and \ref{fig:autonomous_suction}. The suction tool, utilizing the learned PSM Reach policy, reached the threshold distance of 3 mm for every goal and removed the fake blood in both experiments. For the autonomous debris removal, the learned PSM Pick policy on the LND successfully grasped all the debris and reached the threshold distance of 3mm. The learned PSM Reach policy on the LND also successfully handed all the debris to the first assistant and reached the threshold distance.  

\section{Discussion and Conclusion}
In this work, we present the first, open-sourced RL environment for surgical robotics called dVRL. dVRL provides a syntatically common RL environments to OpenAI Gym with a simulation of the da Vinci\textregistered{} Surgical Robot system, a widely used platform with an international network of academic research platforms for which to transfer learned policies onto a real robot environment. 
Using  state-of-art techniques from the RL community such as DDPG and HER, we show that through dVRL control policies were effectively learned and, importantly, could be transferred effectively to a real robot with minimal effort.
While the proposed environments result in simple primitives, reaching and picking, we still showed their utility in a realistic surgical setting via suctioning and debris removal. 
We see dVRL as enabling the broad surgical robotics community to fully leverage the newest strategies in reinforcement learning, and for reinforcement learning scientists with no previous domain knowledge of surgical robotics to be able to test and develop new algorithms that can have real-world, positive impact to patient care and the future of autonomous surgery.

Under dVRL, many options exist moving forward. 
First, the simulator allows for easy additions of new rigid objects, such as needles, to learn more advanced control policies. 
Modeling of endoscopic stereo cameras with their uniquely tight disparities and narrow field of view would allow for visual servoing and visuo-motor policy approaches to be explored. 
Promising future extensions for dVRL to address new applications via packages defining soft body tissue interactions, as demonstrated via Bullet integration \cite{matas2018sim}, thread simulation as demonstrated by Tang et al. \cite{thread_simulation},  rigid tissue interactions such as with bone \cite{mohamed2018modular} and cartilage, and fluid simulation via NVIDIA FleX \cite{nvidia_flex}.

\section{Acknowledgements}
The authors were supported on an 2018 Intuitive Surgical Technology Grant, and would like to thank Dale Bergman, Simon Dimaio, and Omid Mohareri for their assistance with the dVRK.

\balance
\bibliographystyle{ieeetr}
\bibliography{references}
\end{document}